%% file: main.tex
\documentclass{article}

\usepackage{microtype}
\usepackage{graphicx}
\usepackage{subfigure}
\usepackage{booktabs} 
\usepackage{multirow}
\usepackage{amsmath}
\usepackage{hyperref}
\usepackage{soul}
\usepackage{xcolor}
\usepackage{mathtools}
\usepackage{amsmath}

\newcommand{\citeay}[1]{\citeauthor{#1}, \citeyear{#1}}

\usepackage[accepted]{icml2020}

\icmltitlerunning{Aligned Cross Entropy for Non-Autoregressive Machine Translation}

\begin{document}

\twocolumn[
\icmltitle{Aligned Cross Entropy for Non-Autoregressive Machine Translation}




\begin{icmlauthorlist}
\icmlauthor{Marjan Ghazvininejad}{fair}
\icmlauthor{Vladimir Karpukhin}{fair}
\icmlauthor{Luke Zettlemoyer}{fair}
\icmlauthor{Omer Levy}{fair}
\end{icmlauthorlist}

\icmlaffiliation{fair}{Facebook AI Research}

\icmlcorrespondingauthor{Marjan Ghazvininejad}{ghazvini@fb.com}

\icmlkeywords{Machine Translation, Cross Entropy, Non-Autoregressive, Natural Language Processing, Transformer}

\vskip 0.3in
]


\printAffiliationsAndNotice{}  

\input{00-abstract}
\input{01-intro.tex}
\input{02-method.tex}

\input{03-training.tex}

\input{04-experiments.tex}

\input{05-analysis.tex}

\input{06-related.tex}

\input{07-conclusion.tex}

\section*{Acknowledgements}
We thank Abdelrahman Mohamed for sharing his expertise on non-autoregressive models,
and our colleagues at FAIR for valuable feedback.

\bibliography{references}
\bibliographystyle{icml2020}

\end{document}

%% file: 00-abstract.tex
\begin{abstract}
Non-autoregressive machine translation models significantly speed up decoding by allowing for parallel prediction of the entire target sequence. However, modeling word order is more challenging due to the lack of autoregressive factors in the model. This difficultly is compounded during training with cross entropy loss, which can highly penalize small shifts in word order. In this paper, we propose aligned cross entropy (AXE) as an alternative loss function for training of non-autoregressive models. AXE uses a differentiable dynamic program to assign loss based on the best possible monotonic alignment between target tokens and model predictions. AXE-based training of conditional masked language models (CMLMs) substantially improves performance on major WMT benchmarks, while setting a new state of the art for non-autoregressive models.



\end{abstract}

%% file: 01-intro.tex
\section{Introduction}

Non-autoregressive machine translation models can significantly improve decoding speed by predicting every word in parallel~\cite{gu2017,libovicky2018}.
This advantage comes at a cost to performance since modeling word order is trickier when the model cannot condition on its previous predictions.
A range of semi-autoregressive models \cite{lee2018,stern2019,gu2019levenshtein,ghazvininejad2019} have shown there is a speed-accuracy tradeoff that can be optimized with limited forms of autoregression. However, increasing performance of the purely non-autoregressive models without sacrificing decoding speed remains an open challenge.  
In this paper, we present a new training loss for non-autoregressive machine translation that softens the penalty for word order errors, and significantly improves performance with no modification to the model or to the decoding algorithm.

\begin{figure}[t]
\centering
\small
\begin{tabular}{@{}clllll@{}}
\toprule
\textbf{Target $Y$} & \textit{it} & \textit{tastes} & \textit{pretty} & \textit{good} & \textit{though} \\
\midrule
\multirow{5}{*}{\parbox{1.52cm}{\centering \textbf{Model\\Predictions $P$ (Top 5)}}}  
& but & it  & tastes & delicious & . \\
& however & that & makes  & \hl{good} & , \\
& for & this & looks & tasty & so \\
& and & for & taste & fine & \hl{though} \\
& though &  the & feels & exquisite & ! \\
\bottomrule
\end{tabular}

\caption{The model predictions are quite similar to the target, but misaligned by one token. The first and second target tokens (\textit{it tastes}) are predicted in the second and third positions, respectively, leaving only the predictions in the fourth and fifth positions aligned with the target. 
The cross-entropy loss will heavily penalize the predictions in the first, second, and third positions.}
\label{fig:intro}
\end{figure} 

Existing models (both autoregressive and non-autoregressive) are typically trained with cross entropy loss.
Cross entropy is a strict loss function, where a penalty is incurred for every word that is predicted out of position, even for output sequences with small edit distances (see Figure~\ref{fig:intro}).
Autoregressive models learn to avoid such penalties, since words are generated conditioned on the sentence prefix. However, non-autoregressive models do not know the exact sentence prefix, and should (intuitively) focus more on root errors (e.g. a missing word) while allowing more partial credit for cascading errors (the right word in the wrong place). 



To achieve this more relaxed loss, we introduce aligned cross entropy (AXE), a new objective function that computes the cross entropy loss based on an alignment between the sequence of token labels and the sequence of token distribution predictions.
AXE uses dynamic programming to find the monotonic alignment that minimizes the cross entropy loss.
It provides non-autoregressive models with a more accurate training signal by ignoring absolute positions and focusing on relative order and lexical matching. 
We efficiently implement AXE via matrix operations, and use it to train conditional masked language models (CMLM; \citeay{ghazvininejad2019}) for machine translation.
AXE only slightly increases training time compared to cross entropy, and requires no changes to parallel argmax decoding.

Extensive experiments on machine translation benchmarks demonstrate that AXE  substantially boosts the performance of CMLMs, while having the same decoding speed.
In WMT'14 EN-DE, training CMLMs with AXE (instead of the regular cross entropy loss) increases performance by 5 BLEU points; we observe similar trends in WMT'16 EN-RO and WMT'17 EN-ZH.
Moreover, AXE CMLMs significantly outperform state-of-the-art non-autoregressive models, such as FlowSeq \cite{ma2019flowseq}, as well as the recent CRF-based \emph{semi-autoregressive} model with bigram LM decoding \cite{sun2019}.
Our detailed analysis suggests that training with AXE makes models more confident in their predictions, thus reducing multimodality, and alleviating a key problem in non-autoregressive machine translation.

%% file: 02-method.tex
\section{Aligned Cross Entropy}

Let $Y$ be a target sequence of $n$ tokens $Y_1 , \ldots , Y_n$, and $P$ be the model predictions, a sequence of $m$ token probability distributions $P_1 , \ldots , P_m$.
Our goal is to find a monotonic alignment between $Y$ and $P$ that will minimize the cross entropy loss, and thus focus the penalty on lexical errors (predicting the wrong token) rather than positional errors (predicting the right token in the wrong place).

We define an alignment $\alpha$ to be a function that maps target positions to prediction positions, i.e. $\alpha : \{1,\ldots ,n\} \rightarrow \{1,\ldots ,m\}$.  We further assume that this alignment is monotonic, i.e.  $i \leq j$ iff $\alpha(i)\leq \alpha(j)$.
Given a specific alignment $\alpha$, we define a conditional $AXE$ loss as: 

\begin{equation}
\begin{aligned}
\displaystyle
    AXE(&Y_1 , \ldots , Y_n,  P_1 , \ldots , P_m | \alpha) = \\  
    & - \sum_{i=1}^n \log P_{\alpha(i)}(Y_i)     \\ 
    & - \hspace{-35pt} \sum_{k\in \{1 \ldots m\} \setminus \{\alpha(1), \ldots \alpha(n)\}}  \hspace{-35pt} \log P_k (\varepsilon) 
\end{aligned}
\end{equation}

\input{02B-algorithm.tex}
The first term of this loss function is an aligned cross entropy between $Y$ and $P$, and the second term is a penalty for unaligned predictions. Epsilon ($\varepsilon$) is a special ``blank'' token in our vocabulary that appears in the probability distributions, but that does not appear in the final output string. 

Now, the final $AXE$ loss is the minimum over all possible monotonic alignments of the conditional loss:

\begin{equation}
\label{eq:loss}
\begin{aligned}
\displaystyle
  &AXE(Y_1 , \ldots , Y_n,  P_1 , \ldots , P_m) = \\ \\
      &\min _{\alpha }\hspace{2pt} AXE(Y_1 , \ldots , Y_n,  P_1 , \ldots , P_m | \alpha) =\\ 
    &  \min_{\alpha(1) \ldots \alpha(n)} \left(- \sum_{i=1}^n  \log P_{\alpha(i)}(Y_i)   \right. \\
    & \hspace{47pt} \left. - \hspace{-35pt} \sum_{k\in \{1 \ldots m\} \setminus \{\alpha(1), \ldots \alpha(n)\}}  \hspace{-35pt} \log P_k (\varepsilon) \right) \\ \\
    & \mbox{s.t.  } 1 \leq \alpha(1) \leq \alpha(2) \leq \alpha(3) \ldots \leq \alpha(n) \leq m
\end{aligned}
\end{equation} 

Finding the optimal monotonic alignment between two sequences is a well studied problem. For instance, dynamic time warping (DTW) \cite{dtw} is a well-known algorithm for finding the optimal alignment between two different time series. Here we have extended the idea to compute the optimal alignment between a sequence of target tokens and a sequence of prediction probability distributions. We use a simple dynamic program to find the optimal alignment while calculating the AXE loss.

\input{02A-operators.tex}

\begin{figure*}[t!]
\centering
\begin{tabular}{clllll}
\toprule
\textbf{Target $Y$} & \textit{it} & \textit{tastes} & \textit{pretty} & \textit{good} & \textit{though} \\
\midrule
\textbf{Alignment $\alpha: Y \rightarrow P$} 
& 2 & 3 & 3 & 4 & 5 \\
\midrule 
\multirow{5}{*}{\parbox{4cm}{\centering \textbf{Model Predictions $P$}\\\textbf{(Top 5)}}}  
& \multicolumn{1}{l}{but} & \multicolumn{1}{l}{{\hl{it}}}  & \multicolumn{1}{l}{{\hl{tastes}}} & \multicolumn{1}{l}{delicious}  & $\varepsilon$ \\
& \multicolumn{1}{l}{however}& \multicolumn{1}{l}{$\varepsilon$} & \multicolumn{1}{l}{makes}  & \multicolumn{1}{l}{\hl{good}} & .  \\
& \multicolumn{1}{l}{{\hl{$\varepsilon$}}} & \multicolumn{1}{l}{that} & \multicolumn{1}{l}{looks} & \multicolumn{1}{l}{tasty} & , \\
& \multicolumn{1}{l}{for} & \multicolumn{1}{l}{this} & \multicolumn{1}{l}{taste} &  \multicolumn{1}{l}{fine} & so\\
& \multicolumn{1}{l}{and} &  \multicolumn{1}{l}{for} & \multicolumn{1}{l}{feels} & \multicolumn{1}{l}{exquisite} & \hl{though}  \\
\bottomrule
\end{tabular}
\caption{An example illustrating how AXE aligns model predictions with the target sequence. 
The operations that created the optimal alignment in this example are: (1) skip prediction $P_1$ (cost: $\log P_1(\varepsilon)$),
(2) align $Y_1$ to $P_2$ (cost: $\log P_2(\text{it})$),
(3) align $Y_2$ to $P_3$ (cost: $\log P_3(\text{tastes})$),
(4) skip target $Y_3$ (cost: $\delta\log P_3(\text{pretty})$), 
(5) align $Y_4$ to $P_4$ (cost: $\log P_4(\text{good})$),
(6) align $Y_5$ to $P_5$ (cost: $\log P_5(\text{though})$).
}
\label{fig:example}
\end{figure*}

\paragraph{Dynamic Programming}

Given a sequence of target tokens $Y=Y_1\ldots Y_n$ and a sequence of predictions $P=P_1\ldots P_m$ we propose a method to find the score of the optimal alignment between any prefix of these two sequences $Y_{1:i}=Y_1\ldots Y_i$ and $P_{1:j}=P_1\ldots P_j$, for any $i$ and $j$. The score of the optimal alignment for the full sequences is obtained at $i=n$ and $j=m$.

We start by defining a matrix $A$ of $n+1$ by $m+1$ dimensions, respectively corresponding to $Y$ and $P$, where $A_{i,j}$ represents the minimum loss value for aligning $Y_{1:i}$ to $P_{1:j}$ as defined in Equation~\ref{eq:loss}. We initialize $A_{0,0}$ to be $0$ and then proceed to fill the matrix by taking the local minimum at each cell $A_{i,j}$ from three possible operators: \emph{Align}, \emph{Skip Prediction} , and \emph{Skip Target}.
Table~\ref{tab:operators} describes each operation and its update formula. Once the matrix is full, the cell $A_{n,m}$ will contain the cross entropy loss of the optimal alignment.
Algorithm~\ref{alg:dp} lays out a straightforward implementation of AXE's dynamic program.

According to Equation~\ref{eq:loss}, the optimal alignment can be many-to-one, where multiple target positions can be mapped to a single prediction.  This would be computed by \emph{aligning} the first mapped token and \emph{skipping} the rest of target tokens. To discourage skipping too many target tokens, we penalize skip target operators separately with a parameter $\delta$ as described in Table~\ref{tab:operators}. Setting $\delta=1$ will result in the loss function defined in Equation~\ref{eq:loss}, but as we show in our ablation study (Section~\ref{sec:ablations}), higher $\delta$ values yield better performance in practice. 



\paragraph{Efficient Implementation}

The implementation in Algorithm~\ref{alg:dp} has $O(n \cdot m)$ time complexity.
However, multiple updates of the matrix $A$ can be parallelized on GPUs and other tensor-processing architectures.
Rather than iterating over each cell, we iterate over each \emph{anti-diagonal}, computing all the values along the anti-diagonal in parallel.
In other words, we first compute the values of $[A_{0,1}, A_{1,0}]$, followed by $[A_{0,2}, A_{1,1}, A_{2,0}]$, etc.
Since the number of anti-diagonals is $n + m + 1$, we arrive at a time complexity of $O(n+m)$. 
Since $m$ is typically on the same order of magnitude as $n$, the linear cost of computing AXE during training becomes negligible compared to forward and backward passes through the model.\footnote{Batch implementation of this algorithm is straightforward. By doing so, we are able to achieve training times similar  to (about 1.2 times slower than) training with cross entropy loss. }

\paragraph{Example}

Figure~\ref{fig:example} depicts an example application of AXE.
We see that the predictions are generally good, but start with a shift with respect to the target. 
This misalignment would cause the regular cross entropy loss to severely penalize the first three predictions, even though $P_2$ and $P_3$ are correct when aligned with $Y_1$ and $Y_2$.
AXE, on the other hand, finds an alignment between the target and the predictions, which allows it to focus the penalty on the redundant prediction in $P_1$ and the missing token $Y_3$, i.e. the root errors.


%% file: 02B-algorithm.tex
\begin{algorithm}[tb]
   \caption{Aligned Cross Entropy}
   \label{alg:dp}
\begin{algorithmic}
   \STATE {\bfseries Input:} tokens $Y$, predictions $P$
   \STATE $A_{0,0} = 0$
   \FOR{$i=1$ {\bfseries to} $n$}
   \STATE $A_{i,0} = A_{i-1,0} - \delta \cdot \log P_1 (Y_i)$
   \ENDFOR
   \FOR{$j=1$ {\bfseries to} $m$}
   \STATE $A_{0,j} = A_{0,j-1} - \log P_j (\varepsilon)$
   \ENDFOR
   \FOR{$i=1$ {\bfseries to} $n$}
   \FOR{$j=1$ {\bfseries to} $m$}
   \STATE $\text{align} = A_{i-1,j-1} - \log P_j (Y_i)$
   \STATE $\text{skip\_prediction} = A_{i,j-1} - \log P_j (\varepsilon)$
   \STATE $\text{skip\_target} = A_{i-1,j} - \delta \cdot \log P_j (Y_i)$
   \STATE $A_{i,j} = \min \{ \text{align}, \text{skip\_prediction}, \text{skip\_target} \}$
   \ENDFOR
   \ENDFOR
   \STATE {\bfseries return} $A_{n,m}$
\end{algorithmic}
\end{algorithm}

%% file: 02A-operators.tex
\begin{table*}[ht!]
\centering
\begin{tabular}{l p{250pt} l}
\toprule


\textbf{Align} & 
Aligns the current target $Y_i$ with the current prediction $P_j$, updating $A$ along the diagonal. 
& $A_{i,j} = A_{i-1,j-1} - \log P_j (Y_i)$ \\
& & \\

\textbf{Skip Prediction} & 
Skips the current prediction $P_j$ by predicting an empty token ($\varepsilon$), updating $A$ along the $P$ axis. This operation is akin to inserting an empty token to the target sequence at the $i$-th position. 
& $A_{i,j} = A_{i,j-1} - \log P_j (\varepsilon)$ \\
& & \\

\textbf{Skip Target} & 
Skips the current target $Y_i$ by predicting it without incrementing the prediction iterator $j$, updating $A$ along the $Y$ axis. This operation is akin to duplicating the prediction $P_j$. The hyperparameter $\delta$ controls how expensive this operation is; high values of $\delta$ will discourage alignments that skip too many target tokens. 
& $A_{i,j} = A_{i-1,j} - \delta \cdot \log P_j (Y_i)$ \\

\bottomrule
\end{tabular}
\caption{The three local update operators in AXE's dynamic program.}
\label{tab:operators}
\end{table*}

%% file: 03-training.tex
\section{Training Non-Autoregressive Models}
\label{sec:cmlm}

We use AXE to train conditional masked language models (CMLMs) for non-autoregressive machine translation \cite{ghazvininejad2019}.\footnote{While in this work we apply AXE to CMLMs, the loss function can be used to train other models as well. We leave further investigation of this direction to future work.}


\subsection{Conditional Masked Language Models}
\label{sec:cmlm_background}
A conditional masked language model takes a source sequence $X$ and a partially-observed target sequence $Y_{\text{obs}}$ as input, and predicts the probabilities of the masked (unobserved) target sequence tokens $Y_{\text{mask}}$. 
The underlying architecture is an encoder-decoder transformer \cite{vaswani2017}.

In the original paper, CMLMs are used for machine translation where 
a random subset of $Y$ tokens are masked at training time. However, at inference all target tokens are masked ($Y = Y_{\text{mask}}$) and the length of $Y$ (the number of masked tokens) is unknown. To estimate the length of $Y$, an auxiliary task is introduced to predict the target length based on the source sequence $X$.\footnote{See \cite{ghazvininejad2019} for further detail.}

\subsection{Adapting CMLMs to AXE}
\label{sec:cmlm_changes}

In our case, the model can also produce blank tokens ($\varepsilon$), which effectively shorten the predicted sequence's length.
To account for potentially skipped tokens during inference, we multiply the predicted length by a hyperparameter $\lambda$ (which is tuned on the validation set) before applying argmax decoding.

\subsection{Adapting the Training Objectives to AXE}
\label{sec:training_objectives}

Since this work focuses on the purely non-autoregressive setting, the entire target sequence will be masked at inference time ($Y_{\text{mask}} = Y$).
The same does not have to hold for training; we can utilize partially observed sequences in order to provide the learner with easier and more focused training examples.
We experiment with three variations:

\paragraph{Unobserved Input, Predict All}
All the tokens in the target sequence are masked, and the model is expected to predict all of them.
This is a direct replication of the task at inference time.
While AXE allows for the number of masked tokens $m$ to be different from the length of the gold target sequence $n$, we found that setting $m=n$ produced better models in preliminary experiments.

\paragraph{Partially-Observed Input, Predict All}
As in the original CMLM training process, a random subset of the target sequence is masked before being passed onto the model as input.\footnote{The number of masked input tokens is distributed uniformly between $1$ and $n$.}
We then apply AXE on the entire sequence, regardless of which tokens were observed.
When training on partially-observed inputs, we always set $m=n$ to avoid further alterations of the gold target sequence beyond masking.

\paragraph{Partially-Observed Input, Predict Masks}
The straightforward application of AXE to CMLM training (which ignores whether each token was masked or observed) works well in practice.
However, we can also allow AXE to skip the observed tokens when computing cross entropy, and focus the training signal on the actual task.
We do so by setting $P_i (Y_i) = 1$ for every observed token $Y_i$; i.e. if the $i$-th token is observed and is aligned with the prediction corresponding to the same position ($P_i$), there is no penalty.
Our ablation studies show that this modification provides a modest but consistent boost in performance (see Section~\ref{sec:ablations}). As a result, we use this setting for training our model.

%% file: 04-experiments.tex
\begin{table*}[t!]
\centering
\small
\begin{tabular}{lcccccc}
\toprule
\multirow{2}{*}{\textbf{Model}} & \multicolumn{2}{c}{\textbf{WMT'14}} & \multicolumn{2}{c}{\textbf{WMT'16}} & \multicolumn{2}{c}{\textbf{WMT'17}} \\ 
& \textbf{EN-DE} & \textbf{DE-EN} & \textbf{EN-RO} & \textbf{RO-EN} & \textbf{EN-ZH} & \textbf{ZH-EN}  \\\midrule 
\textit{Cross Entropy CMLM  \cite{ghazvininejad2019}} & 18.05 & 21.83 & 27.32 & 28.20 & 24.23 & 13.64 \\
\textit{AXE CMLM (Ours)} & \textbf{23.53} & \textbf{27.90} & \textbf{30.75} & \textbf{31.54} & \textbf{30.88}& \textbf{19.79} \\ 
\bottomrule
\end{tabular}
\caption{The performance (test set BLEU) of AXE CMLM compared to cross entropy CMLM on all of our benchmarks. Both models are purely non-autoregressive, using a single forward pass during argmax decoding.}
\label{tab:xe_results}
\end{table*}

\begin{table*}[t!]
\centering
\small
\begin{tabular}{lrcccc}
\toprule
\multirow{2}{*}{\textbf{Model}} &  \textbf{Decoding} & \multicolumn{2}{c}{\textbf{WMT'14}} & \multicolumn{2}{c}{\textbf{WMT'16}} \\
& \textbf{Iterations} & \textbf{EN-DE} & \textbf{DE-EN} & \textbf{EN-RO} & \textbf{RO-EN} \\\midrule
\textbf{Autoregressive} &&&&& \\
\midrule
\textit{Transformer Base} & $N$~~~~~~& 27.61 & 31.38 & 34.28 &  33.99 \\
\textit{~~~~+ Knowledge Distillation} & $N$~~~~~~& 27.75 & 31.30 & ---~--- &  ---~--- \\
\midrule
\textbf{Non-Autoregressive} &&&&&  \\
\midrule
\textit{Iterative Refinement \cite{lee2018}} &  1~~~~~~& 13.91 & 16.77& 24.45 & 25.73\\
\textit{CTC Loss \cite{libovicky2018} }  & 1~~~~~~& 17.68 & 19.80 & 19.93 & 24.71 \\
\textit{NAT w/ Fertility \cite{gu2017} } & 1~~~~~~& 17.69 & 21.47 & 27.29 &  29.06\\
\textit{Cross Entropy CMLM \cite{ghazvininejad2019}} & 1~~~~~~&18.05 & 21.83 & 27.32 & 28.20 \\
\textit{Auxiliary Regularization \cite{Wang2019NonAutoregressiveMT}} & 1~~~~~~& 20.65 &24.77 &---~--- & ---~--- \\
\textit{Bag-of-ngrams Loss \cite{shao2019minimizing}} & 1~~~~~~& 20.90& 24.61&28.31 &29.29 \\ 
\textit{Hint-based Training \cite{li2019}} & 1~~~~~~& 21.11& 25.24&---~--- &---~--- \\ 
\textit{FlowSeq \cite{ma2019flowseq}} & 1~~~~~~&  21.45 & 26.16   & 29.34 & 30.44 \\
\textit{Bigram CRF \cite{sun2019}} & 1~~~~~~& 23.44 &  27.22 &  ---~--- & ---~---\\
\textit{AXE CMLM (Ours)} & 1~~~~~~&  \textbf{23.53}   & \textbf{27.90} & \textbf{30.75} & \textbf{31.54} \\

\bottomrule
\end{tabular}
\caption{The performance (test set BLEU) of CMLMs trained with AXE, compared to other non-autoregressive methods. The standard (autoregressive) transformer results are also reported for reference.}
\label{tab:main_results}
\end{table*}

\section{Experiments}

We evaluate CMLMs trained with AXE on 6 standard machine translation benchmarks, and demonstrate that AXE significantly improves performance over cross entropy trained CMLMs and over recently-proposed non-autoregressive models as well.

\subsection{Setup}

\paragraph{Translation Benchmarks}
We evaluate our method on both directions of three standard machine translation datasets with various training data sizes:  WMT'14 English-German (4.5M sentence pairs),  WMT'16  English-Romanian (610k pairs), and WMT'17 English-Chinese (20M pairs). The datasets are tokenized into subword units using BPE \cite{sennrich2016}.\footnote{We run joint BPE for all language pairs except English-Chinese.} 
We use the same data and preprocessing as \citet{vaswani2017}, \citet{lee2018}, and \citet{wu2019pay} for WMT'14 EN-DE, WMT'16 EN-RO, and WMT'17 EN-ZH respectively.
We evaluate performance with BLEU \cite{papineni2002} for all language pairs, except for translating English to Chinese where we use SacreBLEU \cite{post2018}.\footnote{SacreBLEU hash: BLEU+case.mixed+lang.en-zh +numrefs.1+smooth.exp+test.wmt17+tok.zh+version.1.3.7}

\paragraph{Hyperparameters}
We generally follow the transformer \emph{base} hyperparameters \cite{vaswani2017}: 6 layers for the encoder and decoder, 8 attention heads per layer,  512 model dimensions, and 2048 hidden dimensions. We follow the weight initialization schema from BERT \cite{devlin2018}, and sample weights from  $\mathcal{N}(0, 0.02)$, set biases to zero, and set layer normalization parameters to $\beta=0$ and $\gamma=1$. For regularization, we set dropout to $0.3$, and use $0.01$ $L_2$ weight decay and label smoothing with $\varepsilon=0.1$.
We train batches of  128k tokens using Adam \cite{Kingma2015} with $\beta=(0.9, 0.999)$ and $\varepsilon=10^{-6}$. The learning rate warms up to $5\cdot10^{-4}$ within 10k steps, and then decays with the inverse square-root schedule. We train all models for 300k steps. We measure the validation loss at the end of each epoch, and average the 5 best checkpoints based on their validation loss to create the final model. We train all models with mixed precision  floating point arithmetic on 16 Nvidia V100 GPUs. 
For autoregressive decoding, we use a beam size of $b=5$ \cite{vaswani2017} and tune the length penalty on the validation set. Similarly we use $\ell=5$ length candidates for CMLM models, tune the length multiplier ($\lambda \in \{1.05 \ldots 1.1\})$,\footnote{Our preliminary analysis shows that AXE selects Skip Prediction in $5-10\%$ of the time, roughly suggesting that five to ten percent of generated tokens are epsilons. Hence, we search the same range for the length multiplier.} and the target skipping penalty  ($\delta \in \{ 1 \ldots 5 \}$) on the validation set. 

\paragraph{Knowledge Distillation}
Similar to previous work on non-autoregressive translation \cite{gu2017,lee2018, ghazvininejad2019,stern2019}, we  use sequence-level knowledge distillation~\cite{Kim2016SequenceLevelKD} by training CMLMs on translations generated by a standard left-to-right transformer model (transformer \emph{large} for WMT'14 EN-DE and WMT'17 EN-ZH, transformer \emph{base} for WMT'16 EN-RO).
We report the performance of standard autoregressive base transformers trained on distilled data for WMT'14 EN-DE and WMT'17 EN-ZH.

\begin{table*}[t!]
\centering
\small
\begin{tabular}{lrcccc}
\toprule
\multirow{2}{*}{\textbf{Model}} &  \textbf{Decoding} & \multicolumn{2}{c}{\textbf{WMT'14}} & \multicolumn{2}{c}{\textbf{WMT'16}} \\
& \textbf{Iterations} & \textbf{EN-DE} & \textbf{DE-EN} & \textbf{EN-RO} & \textbf{RO-EN} \\\midrule
\textbf{Knowledge Distillation} &&&&& \\ \midrule
\textit{AXE CMLM} (Ours) & 1~~~~~~&  23.53    & 27.90 & 30.75 & 31.54 \\ \midrule
\textbf{Raw Data} &&&&& \\ \midrule
\textit{Cross Entropy CMLM  \cite{ghazvininejad2019}} & 1~~~~~~& 10.64 &  ---~---  &21.22  &  ---~--- \\
\textit{CTC Loss \cite{libovicky2018} }  & 1~~~~~~& 17.68 & 19.80 & 19.93 & 24.71 \\
\textit{FlowSeq \cite{ma2019flowseq}} & 1~~~~~~&   18.55& 23.36& 29.26 &30.16 \\
\textit{AXE CMLM} (Ours) & 1~~~~~~& \textbf{20.40} & \textbf{24.90} & \textbf{30.47} & \textbf{31.42}\\ 
\bottomrule
\end{tabular}
\caption{The performance (test set BLEU) of AXE CMLM, compared to other non-autoregressive methods on raw data. The result of AXE CMLM trained with distillation is also reported as a reference.}
\label{tab:raw}
\end{table*}

\subsection{Main Results}

\paragraph{AXE vs Cross Entropy}
We first compare the performance of AXE-trained CMLMs to that of CMLMs trained with the original cross entropy loss.
Table~\ref{tab:xe_results} shows that training with AXE substantially increases the performance CMLMs across all benchmarks. On average, we gain 5.2 BLEU by replacing cross entropy with AXE, with gains of up to 6.65 BLEU in WMT'17 EN-ZH.

\paragraph{State of the Art}
We compare the performance of CMLMs with AXE against nine strong baseline models: the fertility-based sequence-to-sequence model \cite{gu2017}, transformers trained with CTC loss \cite{libovicky2018}, the iterative refinement approach \cite{lee2018}, transformers trained with auxiliary regularization \cite{Wang2019NonAutoregressiveMT}, CMLMs trained with (regular) cross entropy loss \cite{ghazvininejad2019}, Flowseq: a latent variable model based on generative flow \cite{ma2019flowseq}, hint-based training \cite{li2019}, bag-of-ngrams training \cite{shao2019minimizing}, and the CRF-based semi-autoregressive model \cite{sun2019}. All of these models except the last one are purely non-autoregressive, while the CRF-based model uses bigram statistics during decoding, which deviates from the purely non-autoregressive setting.\footnote{CMLMs \cite{ghazvininejad2019} and the iterative refinement method \cite{lee2018} are presented as semi-autoregressive models that run in multiple decoding iterations. However, the first decoding iteration of these models is purely non-autoregressive, which is what we use as our baselines.}

Table~\ref{tab:main_results} shows that our system yields the highest BLEU scores of all non-autoregressive models. AXE-trained CMLMs outperform the best purely non-autoregressive model (FlowSeq) on both directions of WMT'14 EN-DE and WMT'16 EN-RO by 1.6 BLEU on average. Moreover, our approach achieves higher BLEU scores than the semi-autoregressive CRF decoder across all available benchmarks.

\paragraph{Raw Data}
Finally, we compare the performance of AXE to other methods that train on raw data \emph{without} knowledge distillation.
Table~\ref{tab:raw} shows that AXE CMLMs still significantly outperform other non-autoregressive models in the raw data scenario. In addition, comparing raw data to knowledge distillation training follows previously-published results that demonstrate the importance of knowledge distillation for non-autoregressive approaches ~\cite{gu2017,ghazvininejad2019,zhou2019understanding}, although the gap is much smaller for WMT'16 EN-RO.

\subsection{Ablation Study}\label{sec:ablations}

In this section, we consider several variations of our proposed method to investigate the effect of each component. We test the performance of AXE CMLMs with these variations on the WMT'14 DE-EN and EN-DE datasets. To prevent overfitting, we evaluate on the validation set using $\ell = 5$ length candidates.

\begin{table}[t!]
\centering
\small
\begin{tabular}{@{}llll@{}}
\toprule
\multicolumn{2}{c}{\textbf{Training Objective}}  & \multicolumn{2}{c}{\textbf{WMT'14}} \\
\textbf{Input Tokens} & \textbf{Loss Function} & \textbf{EN-DE} & \textbf{DE-EN}\\
\midrule
\textit{Unobserved}& \textit{All Tokens} & 21.97  & 26.32  \\
\textit{Partially-Observed} & \textit{All Tokens}& 22.80 & 27.59   \\
\textit{Partially-Observed} &  \textit{Only Masks}& \textbf{23.13} & \textbf{28.01}   \\
\bottomrule
\end{tabular}
\caption{The effect of different training objectives on performance,  measured on WMT’14 DE-EN and EN-DE (validation set BLEU).}
\label{tab:training_objectives}
\end{table}

\textbf{Different Training Objectives}  Table~\ref{tab:training_objectives} shows the effects of different training objectives (Section~\ref{sec:training_objectives}), in which all or part of the target tokens are masked and the loss function is calculated on all tokens or masked tokens only.
We find that simulating the inference scenario, where all tokens are unobserved, is actually less effective than revealing a subset of the target tokens as input during training. We speculate that partially-observed inputs add easier examples to the training set, allowing for better optimization as in curriculum learning \cite{bengio2009curriculum}.
We also see that including only the masked tokens in the loss function gives us a modest but consistent boost in performance, possibly because the training signal is focused on the actual task.


\begin{table}[t!]
\centering
\small
\begin{tabular}{@{}ccccc@{}}
\toprule
\textbf{Skip Target} & \multicolumn{2}{c}{\textbf{WMT'14 EN-DE}} & \multicolumn{2}{c}{\textbf{WMT'14 DE-EN}} \\
 \textbf{Penalty} &  \textbf{BLEU} & \textbf{Skip Target} &  \textbf{BLEU} & \textbf{Skip Target} \\
\midrule
1 & 22.60 & 17.57\% & 26.84 & 16.87\% \\
2 & 23.01 & 10.91\% & 27.77 & 10.53\% \\
3 & 22.85 & 9.56\% & 27.87 & 9.04\% \\
4 & 22.90 & 8.14\% &  \textbf{28.01}& 7.83\% \\
5 & \textbf{23.13} & 7.40\% & 27.79 & 6.95\% \\

\bottomrule
\end{tabular}
\caption{The effect of changing the skip target penalty coefficient $\delta$ on performance (BLEU) and the percentage of target words that were skipped, using the validation sets of WMT’14 DE-EN and EN-DE.}
\label{tab:skip_target}
\end{table}

\textbf{Skip Target Penalty} 
The hyperparameter $\delta$ acts as a coefficient for the penalty associated with skipping a target token (see Table~\ref{tab:operators} for a definition). 
We experiment with different values of $\delta$, and report our findings in Table~\ref{tab:skip_target}.
We observe that tuning $\delta$ can significantly improve performance with respect to the default of $\delta = 1$.
As intended, high values of $\delta$ discourage alignments that skip too many target tokens.

\textbf{Length Multiplier} 
The length multiplier $\lambda$ inflates the length predicted by a CMLM to account for extra blank tokens ($\varepsilon$) that the model could potentially generate (see Section~\ref{sec:cmlm_changes} for more detail).
Table~\ref{tab:length_multiplier} compares the effect of different length multiplier $\lambda$ values.
Using the best length multiplier increases the performance by 0.53 BLEU on average for WMT'14 EN-DE and WMT'16 EN-RO. 

\begin{table}[t!]
\centering
\small
\begin{tabular}{@{}lllll@{}}
\toprule
\multirow{2}{*}{\textbf{Length Multiplier}} & \multicolumn{2}{c}{\textbf{WMT'14}} & \multicolumn{2}{c}{\textbf{WMT'16}} \\
 &  \textbf{EN-DE} & \textbf{DE-EN} &  \textbf{EN-RO} & \textbf{RO-EN} \\
\midrule
$\lambda=1$ & 22.96& 27.50& 30.43& 32.22\\
$\lambda=1.01$ & 23.06&27.56 & 30.43 &32.25 \\
$\lambda=1.02$ & 23.09& 27.70 & 30.66 &32.50\\
$\lambda=1.03$  & 23.11& 27.81& 30.75 &32.69\\
$\lambda=1.04$  & \textbf{23.13}& 27.85& 30.88 &32.83\\
$\lambda=1.05$ &\textbf{23.13}& 27.93& 30.88&\textbf{32.94} \\
$\lambda=1.06$ & 23.06& \textbf{28.01}& 31.01&32.84\\
$\lambda=1.07$ & 23.09& 27.93& 31.10&32.61\\
$\lambda=1.08$ & 23.06& 27.71& \textbf{31.14}&32.45\\
$\lambda=1.09$ & 23.07& 27.68& 31.06&32.14\\
$\lambda=1.10$ & 22.92 & 27.49& 30.85&32.01\\


\bottomrule
\end{tabular}
\caption{The effect of tuning the length multiplier $\lambda$ on performance (BLEU), using the validation set.}
\label{tab:length_multiplier}
\end{table}

%% file: 05-analysis.tex
\section{Analysis}
We provide a qualitative analysis to provide some insight where AXE improves over cross entropy, and potential directions for future research on non-autoregressive generation. 

\paragraph{AXE Handles Long Sequences Better}
We first measure performance of cross entropy versus AXE-trained CMLMs for different sequence lengths. We use \texttt{compare-mt} \cite{neubig2019} to split the test sets of WMT'14 EN-DE and DE-EN into different buckets based on target sequence length and calculate BLEU for each bucket. Table~\ref{tab:length_vs_bleu} shows that the performance of models trained with cross entropy drops drastically as the sequence length increases, while the performance of AXE-trained models remains relatively stable.
One explanation for this result is that the longer the sequence, the more likely we are to observe misalignments between the model's predictions and the target; AXE realigns these cases, providing the model with a cleaner signal for modeling long sequences.

\begin{table}[t]
\centering
\small
\begin{tabular}{lllcc}
\toprule
&& & \textbf{Cross Entropy}& \textbf{AXE} \\
\midrule
\multirow{12}{*}{{\rotatebox[origin=c]{90}{\textbf{WMT'14}}}} &\multirow{6}{*}{{\rotatebox[origin=c]{90}{\textbf{EN-DE}}}}& $~~1 \leq N < 10$   & 18.75 & 20.48  \\ 
&&$10 \leq N < 20$	&  21.69 & 23.92  \\ 
&&$20 \leq N < 30$	&   18.64 & 24.21  \\ 
&&$30 \leq N < 40$	&   15.37 & 22.65    \\ 
&&$40 \leq N < 50$	&  14.04 & 23.04    \\ 
&&$50 \leq N $	    &  11.62 & 23.43  \\ 
\cmidrule[0.5pt](lr){2-5}
&\multirow{6}{*}{{\rotatebox[origin=c]{90}{\textbf{DE-EN}}}}&$~~1 \leq N < 10$   &22.57&  24.39 \\ 
&&$10 \leq N < 20$	&  25.28 & 27.86\\ 
&&$20 \leq N < 30$	&  22.43 & 28.78\\ 
&&$30 \leq N < 40$	&  19.03 & 27.18  \\ 
&&$40 \leq N < 50$	&  16.16 & 27.55\\ 
&&$50 \leq N $	    &  12.23 & 27.64\\ 
\bottomrule
\end{tabular}
\caption{The performance (test set BLEU) of cross entropy CMLM and AXE CMLM on WMT'14 EN-DE and DE-EN, bucketed by target sequence length ($N$).}
\label{tab:length_vs_bleu}
\end{table}

\begin{figure}[t]
     \centering
     \subfigure[Short sequences (less than 10 tokens).]{\label{fig:hyp_prob:a}\includegraphics[width=0.45\textwidth]{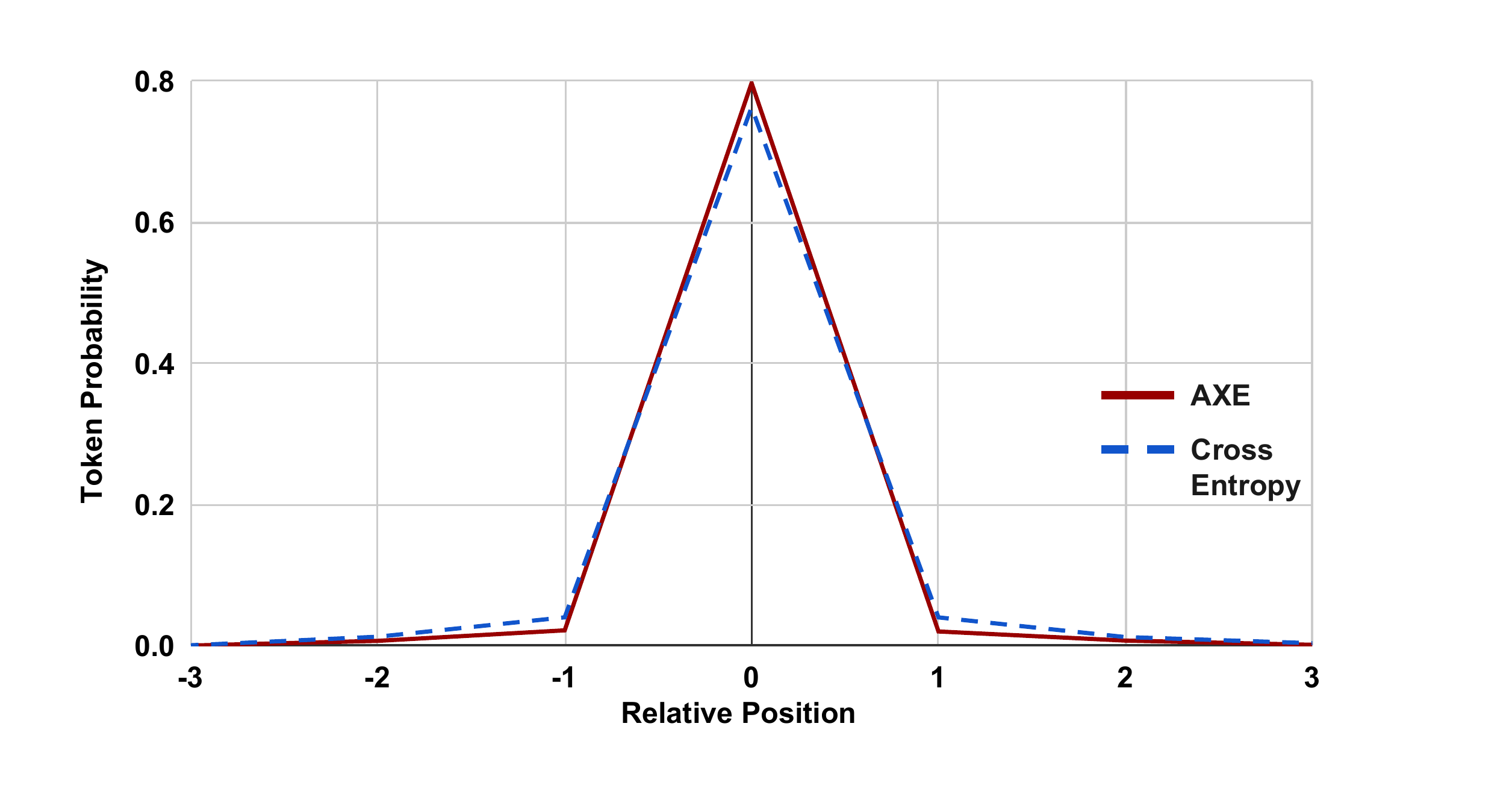}}
      \subfigure[Long sequences (more than 30 tokens).]{\label{fig:hyp_prob:b}\includegraphics[width=0.45\textwidth]{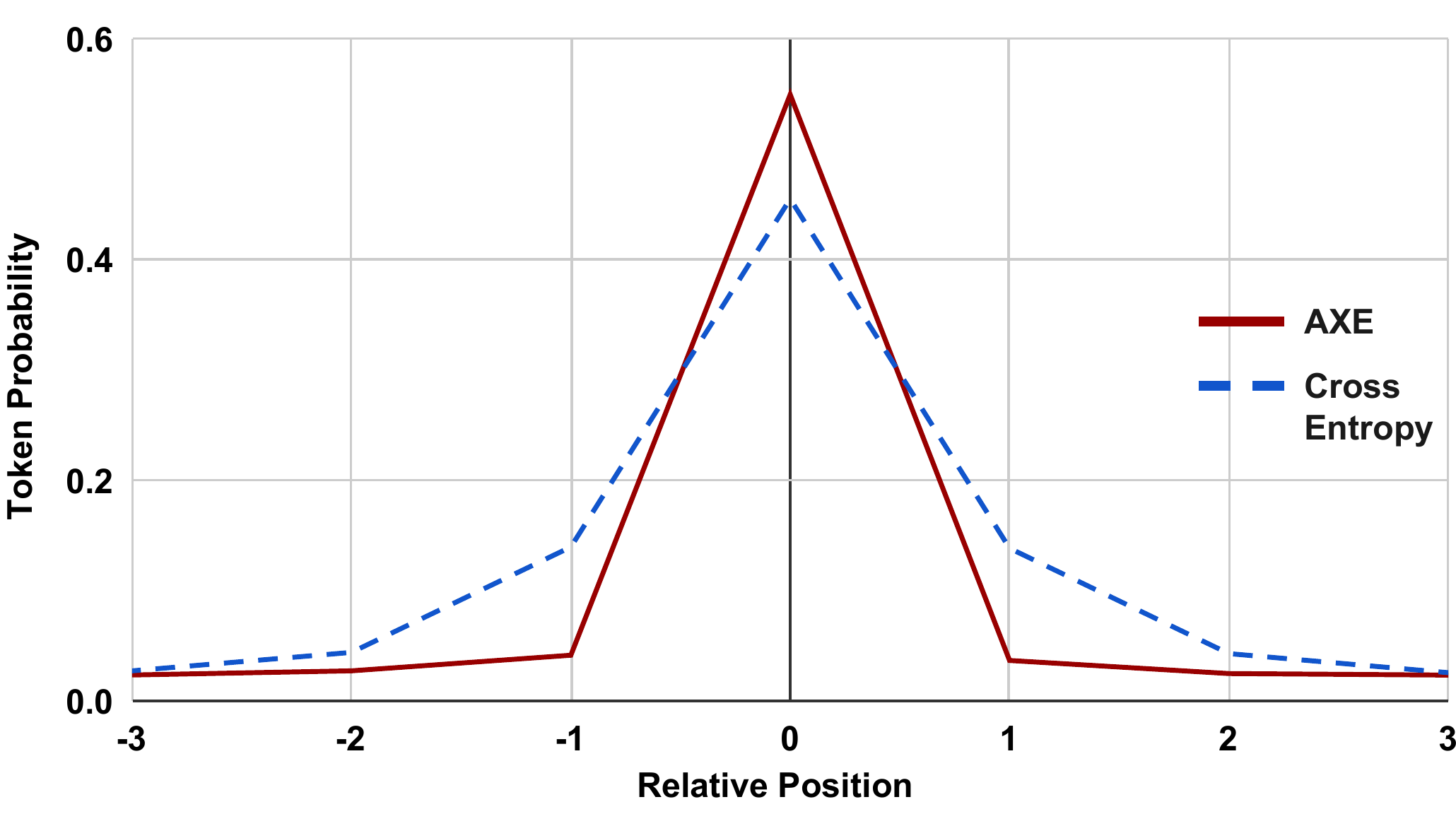}}
     \caption{The average prediction probability assigned to a token as a function of its relative distance from where it was generated in the sequence.
     Each plot shows the average probabilities for CMLMs trained with cross entropy (dashed blue line) and AXE (solid red line).
     }\label{fig:hyp_prob}
\end{figure} 

\paragraph{AXE Increases Position Confidence}
We also study how confident each model is about the position of each generated token.
Ideally, we would like each predicted token to have a high probability at the position in which it was predicted and a very low probability in the neighboring positions. 
After applying argmax decoding, we compute the probability assigned to each generated token in all positions of the sequence and average these probabilities based on the relative distance (positive or negative) to the generated position.
Figure~\ref{fig:hyp_prob} plots these averaged probabilities for both short ($<10$ tokens) and long ($>30$ tokens) target sequences.

Both models are rather confident in their predictions for short sequences (Figure~\ref{fig:hyp_prob:a}): the probability has a high peak at the generated position and drops rapidly as we move further away.
However, for longer sentences (Figure~\ref{fig:hyp_prob:b}), we observe that the plot for cross entropy has lost its sharpness. Specifically, the immediate neighbors of the prediction position ($\pm 1$) receive about $0.14$ probability on average, almost a third of the peak probability. Meanwhile, the probabilities predicted by the AXE-trained model are significantly sharper, assigning negligible probabilities to the generated token in neighboring positions when compared to the center.

On way to explain this result is that cross entropy training encourages predictions to have some probability mass of their neighbors, in order to ``hedge their bets'' in case the predictions are misaligned with the target. Since AXE finds the best alignment before computing the actual loss, spreading the probability mass of a token among its neighbors is no longer necessary.


\paragraph{AXE Reduces Multimodality}
We further argue that AXE reduces the multimodality problem in non-autoregressive machine translation~\cite{gu2017}.
Due to minimal coordination between predictions in many non-autoregressive models, a model might consider many possible translations at the same time. In this situation, the model might merge two or more different translations and generate an inconsistent output that is typically characterized by token repetitions.
We therefore use the frequency of repeated tokens as a proxy for measuring multimodality in a model.

Table~\ref{tab:repetitions} shows the repetition rate for cross entropy and AXE-trained CMLMs. Replacing cross entropy with AXE drastically reduces multimodality, decreasing the number of repetitions by a multiplicative factor of 12.

\begin{table}[t]
\centering
\small
\begin{tabular}{lcc}
\toprule
\multirow{2}{*}{\textbf{Model}} & \multicolumn{2}{c}{\textbf{WMT'14}} \\
& \textbf{EN-DE} & \textbf{DE-EN} \\
\midrule
\textit{Cross Entropy CMLM} & 16.72\% & 12.31\% \\
\textit{AXE CMLM} & ~~1.41\% & ~~1.03\% \\
\bottomrule
\end{tabular}
\caption{The percentage of repeated tokens on the test sets of WMT'14 EN-DE and DE-EN.}
\label{tab:repetitions}
\end{table}

%% file: 06-related.tex
\section{Related Work}

Advances in neural machine translation techniques in recent years has brought an increasing interest in breaking the autoregressive generation bottleneck in translation models.

Semi-autoregressive models introduce partial parallelism into the decoding process. Some of these techniques include iterative refinement of translations based on previous predictions \cite{lee2018,ghazvininejad2019,ghazvininejad2020semi,gu2019levenshtein,kasai2020parallel} and combining a lighter autoregressive decoder with a non-autoregressive one~\cite{sun2019}.

Building a fully non-autoregrssive machine translation model is a much more challenging task. One branch of prior work approaches this problem by modeling with latent variables. \citet{gu2017} introduces word fertility as a latent variable to model the number of generated tokens per each source word.
\citet{ma2019flowseq} uses generative flow to model complex distribution of latent variables for parallel decoding of target.  \citet{shu2019latent} proposes a latent-variable non-autoregressive model with continuous latent variables and a deterministic inference procedure.

There is also work that develops other alternative loss functions for non-autoregressive machine translation. \citet{libovicky2018} use the Connectionist Temporal Classification training objective, a loss function from the speech recognition literature that is designed to eliminating repetitions.
\citet{li2019} uses the learning signal provided by hidden states and attention distributions of an autoregressive teacher. \citet{yang2019non} improves the decoder hidden representations by adding the reconstruction error of source sentence from  these representations as an auxiliary regularization term to the loss function. Finally, \citet{shao2019minimizing} introduce the bag-of-ngrams training objective to encourage the model to capture  target-side sequential dependencies.

%% file: 07-conclusion.tex
\section{Conclusion}
We introduced Aligned Cross Entropy (AXE) as an alternative loss function for training non-autoregressive models. AXE focuses on relative order and lexical matching instead of relying on absolute positions. We showed that, in the context of machine translation, a conditional masked language model (CMLM) trained with AXE significantly outperforms cross entropy trained models, setting a new state-of-the-art for non-autoregressive models.